\PassOptionsToPackage{table}{xcolor}
\documentclass[manuscript,screen]{acmart}

\acmJournal{HEALTH}
\setcopyright{none}
\renewcommand\footnotetextcopyrightpermission[1]{}

\usepackage{algorithm}
\usepackage{algorithmic}
\usepackage{array}
\usepackage{booktabs}
\usepackage{multirow}
\usepackage{tabularx}
\usepackage{float}
\usepackage{placeins}
\usepackage{url}

\definecolor{tabblue}{RGB}{221,235,247}
\definecolor{tabblueDark}{RGB}{0,82,147}
\definecolor{tabgray}{gray}{0.94}
\definecolor{tabgreen}{RGB}{34,139,34}
\definecolor{tabred}{RGB}{185,68,68}
\definecolor{tabmid}{RGB}{100,100,100}
\definecolor{notegold}{RGB}{246,190,0}
\definecolor{TABBLUEDARK}{RGB}{0,82,147}
\definecolor{TABGREEN}{RGB}{34,139,34}
\definecolor{TABRED}{RGB}{185,68,68}
\newcommand{\best}[1]{\textbf{#1}}
\newcommand{\oursmethod}[1]{\textbf{\textcolor{tabblueDark}{#1}}}
\newcommand{\gain}[1]{\textcolor{tabgreen}{\textbf{#1}}}
\newcommand{\drop}[1]{\textcolor{tabred}{\textbf{#1}}}
\newcommand{\neutral}[1]{\textcolor{tabmid}{#1}}
\newcommand{\tabnote}[1]{\textcolor{tabmid}{\footnotesize #1}}

\newcommand{\arxivnotice}{\begingroup\scriptsize\sffamily\itshape\textcolor{tabmid}{\textbf{Notice:} This manuscript is currently under review at \textit{ACM Transactions on Computing for Healthcare} (ACM HEALTH); the final published version shall prevail.}\endgroup}
\makeatletter
\pretocmd{\@mktitle}{\noindent\parbox{\textwidth}{\centering\arxivnotice}\par\vspace{0.8em}}{}{}
\makeatother
\newcolumntype{C}{>{\centering\arraybackslash}c}
\newcolumntype{Y}{>{\centering\arraybackslash}X}

\begin{document}

\title[BMDS-Net: Deployment-aware multi-modal brain tumor segmentation]{BMDS-Net: Deployment-aware multi-modal brain tumor segmentation with adaptive fusion, decoder regularization, and Bayesian calibration}

\author{Yan Zhou}
\authornote{These authors contributed equally to this research.}
\orcid{0009-0004-5324-1112}
\email{202310080133@stu.csust.edu.cn}
\affiliation{%
  \institution{Changsha University of Science and Technology - Yuntang Campus, School of Mathematics and Statistics}
  \city{Changsha}
  \state{Hunan}
  \country{China}}

\author{Suncheng Xiang}
\authornotemark[1]
\orcid{0000-0002-9141-6460}
\affiliation{%
  \institution{Shanghai Jiao Tong University}
  \city{Shanghai}
  \country{China}}

\author{Zhen Huang}
\affiliation{%
  \institution{Changsha University of Science and Technology - Yuntang Campus, School of Mathematics and Statistics}
  \city{Changsha}
  \state{Hunan}
  \country{China}}

\author{Yue Ouyang}
\affiliation{%
  \institution{Changsha University of Science and Technology - Yuntang Campus, School of Mathematics and Statistics}
  \city{Changsha}
  \state{Hunan}
  \country{China}}

\author{Yingqiu Li}
\affiliation{%
  \institution{Changsha University of Science and Technology - Yuntang Campus, School of Mathematics and Statistics}
  \city{Changsha}
  \state{Hunan}
  \country{China}}

\author{Zehua Wang}
\authornote{Corresponding author.}
\email{sjtu\_wzh@sjtu.edu.cn}
\affiliation{%
  \institution{Shanghai Jiao Tong University School of Medicine, Shanghai Chest Hospital}
  \city{Shanghai}
  \state{Shanghai}
  \country{China}}

\begin{abstract}
Multi-modal MRI enables detailed brain tumor sub-region segmentation, but clinical deployment remains affected by missing sequences, boundary errors, and overconfident predictions. We present BMDS-Net, a two-stage Swin UNETR framework that combines adaptive modality fusion, boundary-aware decoder regularization, and efficient Bayesian calibration. The deterministic stage adds zero-initialized multimodal contextual fusion (MMCF) and residual-gated deep decoder supervision (DDS) to modulate modality contributions and regularize hierarchical boundary reconstruction. The second stage replaces only the final convolutional layer with a Bayesian layer, enabling Monte Carlo uncertainty estimates with limited additional training cost.

On BraTS 2021 (1,251 cases), BMDS-Net achieved Dice scores of 0.9293, 0.9098, and 0.8675 for WT, TC, and ET, with HD95 values of 2.27, 2.22, and 3.27~mm. Configuration comparisons showed that auxiliary supervision favored complete-input segmentation, whereas the complete MMCF+gated-DDS configuration was most robust under single-modality removal. Bayesian BMDS-Net achieved an expected calibration error of 0.0037, close to a three-model ensemble (0.0035) at approximately 60\% lower training cost. On BraTS 2020, the same protocol preserved the leading internal ranking of the complete configuration among five Swin UNETR-controlled variants. The implementation code is available at \url{https://github.com/RyanZhou168/BMDS-Net}.
\end{abstract}



\maketitle

\section{Introduction}
\label{sec1}

Gliomas are the most common primary brain tumors, and their management relies heavily on multi-modal magnetic resonance imaging (MRI) for surgical planning, radiation dose contouring, and treatment response assessment \cite{Menze2015BRATS,Bakas2017SciData}. Each MRI sequence contributes distinct diagnostic information: FLAIR highlights peritumoral edema, T1-weighted imaging delineates anatomical boundaries, T1ce reveals enhancing tumor through gadolinium uptake, and T2 provides complementary soft-tissue contrast. Accurate segmentation of the whole tumor (WT), tumor core (TC), and enhancing tumor (ET) from these sequences is essential for quantitative volumetric analysis, yet manual delineation is time-consuming (20--60 minutes per case) and exhibits substantial inter-rater variability \cite{Menze2015BRATS,Baid2021BraTSBenchmark}.

Automated segmentation has progressed rapidly, from CNN-based encoder--decoder architectures such as U-Net \cite{Ronneberger2015UNet}, V-Net \cite{Milletari2016VNet}, and nnU-Net \cite{Isensee2021nnUNet}, to Transformer-based models including UNETR \cite{Hatamizadeh2021UNETR} and Swin UNETR \cite{Hatamizadeh2022SwinUNETR} that capture long-range spatial dependencies through hierarchical self-attention. These advances have pushed benchmark Dice scores above 0.90 for whole-tumor segmentation. However, high benchmark accuracy alone does not guarantee clinical reliability. Two practical gaps remain largely unaddressed in current high-performing architectures:

\textbf{Gap 1: Fragility under incomplete modalities.} In routine clinical practice, acquiring all four MRI sequences is not always feasible. Patients may be unable to tolerate gadolinium contrast (precluding T1ce), scanner time constraints may force protocol abbreviation, or retrospective data may lack specific sequences due to institutional protocol variation \cite{Havaei2016HeMIS}. Standard early-fusion pipelines that concatenate all modalities at the input are inherently brittle in this setting: a zeroed-out channel provides no informative input evidence, and the network has no mechanism to redistribute reliance to the remaining modalities. Dedicated missing-modality methods have addressed this problem in CNN-dominant settings \cite{Havaei2016HeMIS,Neverova2015ModDrop}, leaving their integration with recent Transformer architectures comparatively underdeveloped.

\textbf{Gap 2: Absence of calibrated confidence.} Deterministic segmentation models produce a single prediction without quantifying how confident that prediction is. In clinical decision-making, however, boundary uncertainty directly affects treatment: an uncertain tumor margin may warrant wider radiation margins or additional imaging. Bayesian neural networks \cite{Blundell2015BayesByBackprop}, Monte Carlo dropout \cite{Gal2016DropoutBayesian}, and deep ensembles \cite{Lakshminarayanan2017Ensemble} can provide uncertainty estimates, but their computational cost is prohibitive for large 3D Transformer models. A fully Bayesian Swin UNETR, for instance, would require variational inference over tens of millions of parameters, making training impractical on standard hardware.

Existing literature addresses these gaps largely in isolation. Missing-modality learning focuses on input robustness without considering uncertainty, while Bayesian methods focus on calibration without considering modality degradation. Moreover, decoder-side optimization in Transformer segmentation remains underexplored: most architectural innovation concentrates on the encoder, leaving the decoder with simple upsampling and skip connections that may not fully exploit hierarchical features for boundary-sensitive regions.

To bridge these gaps within a unified framework, we propose \textbf{BMDS-Net} (Bayesian Multi-modal Deep Supervision Network), a two-stage architecture that augments Swin UNETR for reliable multi-modal segmentation.

Our contributions are threefold:
\begin{enumerate}
    \item \textbf{Adaptive fusion}: We introduce zero-initialized MMCF, a modality-fusion module that learns modality-specific attention without perturbing the pretrained Swin UNETR backbone.
    \item \textbf{Boundary-aware decoder regularization}: We design residual-gated DDS, which carries input-level modality saliency into hierarchical decoder stages to regularize boundary-sensitive reconstruction.
    \item \textbf{Efficient calibrated deployment}: We add last-layer Bayesian fine-tuning and evaluate the resulting framework through BraTS 2021 segmentation, missing-modality, and calibration experiments, with BraTS 2020 same-protocol validation of the internal configuration ranking.
\end{enumerate}

The central finding is condition-dependent complementarity. Auxiliary decoder supervision sharpens complete-input boundaries. MMCF stabilizes incomplete-modality inference when paired with gated DDS. Bayesian BMDS-Net provides calibrated confidence summaries without changing the segmentation backbone.

\begin{figure}[!t]
  \centering
  \includegraphics[width=0.9\textwidth]{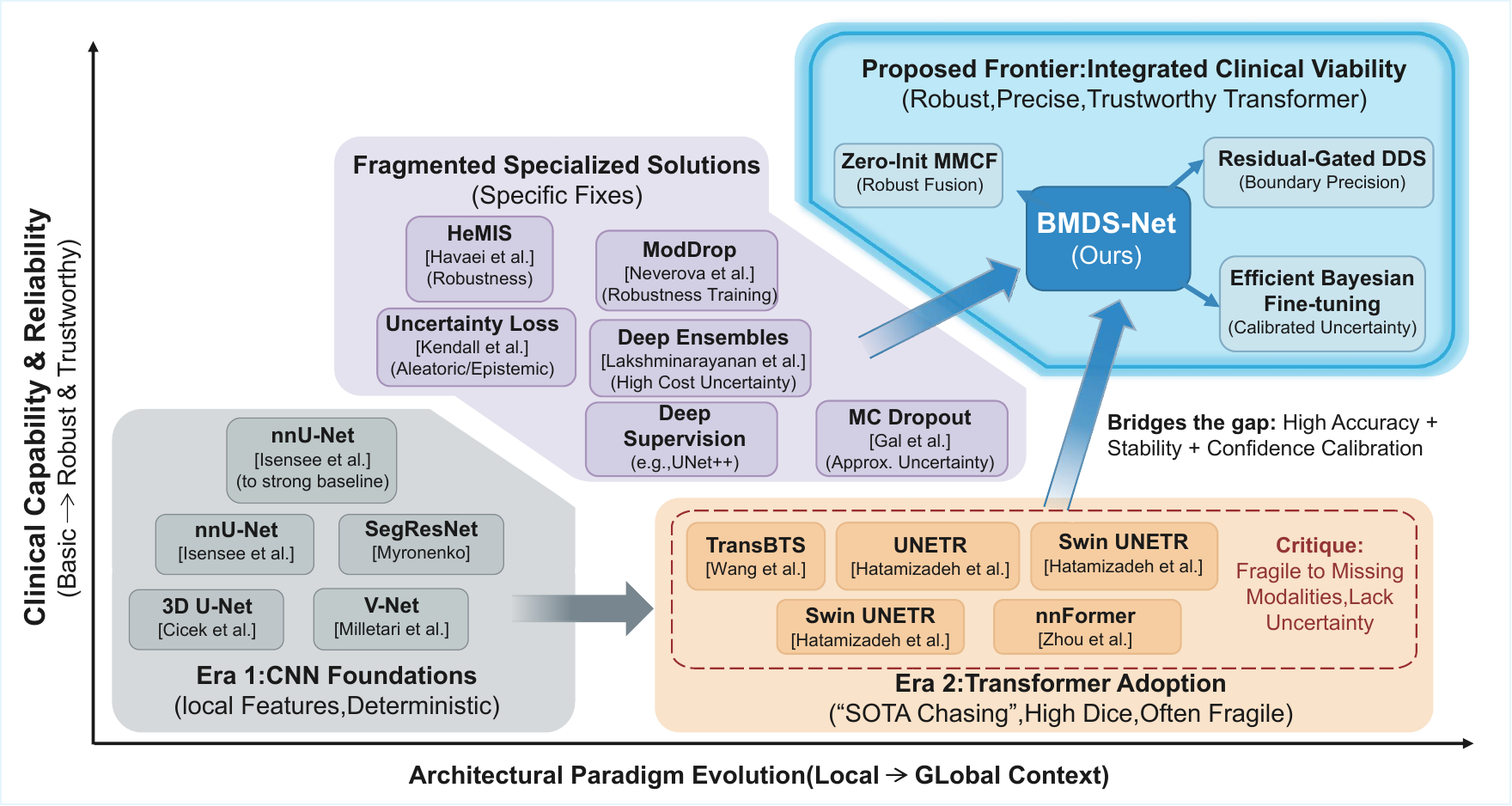}
  \Description{A schematic map positioning BMDS-Net among brain tumor segmentation directions including CNNs, Transformers, robustness, and uncertainty estimation.}
  \caption{Schematic overview of representative research directions in brain tumor segmentation. The horizontal axis illustrates the progression from CNN-based local feature modeling to Transformer-based global context modeling. The vertical axis summarizes practical capabilities relevant to clinical deployment. Existing studies have largely emphasized either benchmark accuracy, robustness to incomplete modalities, or uncertainty estimation in isolation. BMDS-Net integrates these directions within a single Transformer-based framework.}
  \label{fig:lit_review}
\end{figure}

Figure~\ref{fig:lit_review} provides a schematic overview of how BMDS-Net relates to existing research directions in brain tumor segmentation.

The remainder of this paper is organized as follows. Section~\ref{sec2} reviews related work on multi-modal fusion, decoder optimization, and uncertainty estimation. Section~\ref{sec3} details the proposed method. Section~\ref{sec4} presents experiments including comparisons with representative methods, configuration comparisons, BraTS 2020 second-dataset validation, robustness analysis, and uncertainty evaluation. Section~\ref{sec5} discusses the findings, design choices, failure cases, and limitations. Section~\ref{sec6} concludes the paper.

\section{Related Work}
\label{sec2}

\subsection{Multi-modal fusion strategies for brain tumor segmentation}

Brain tumor segmentation exploits complementary information from multiple MRI sequences. The dominant fusion strategy in current architectures is early concatenation, where all modalities are stacked as input channels \cite{Cicek20163DUNet,Wang2021TransBTS,Hatamizadeh2022SwinUNETR}. This approach is simple and effective when all modalities are present, but it treats each modality as equally reliable and provides no mechanism to adapt when a modality is degraded or absent.

To address modality heterogeneity, HeMIS \cite{Havaei2016HeMIS} proposed learning hetero-modal representations by computing statistics (mean and variance) across available modality embeddings, enabling inference with arbitrary subsets. ModDrop \cite{Neverova2015ModDrop} simulated missing modalities during training through stochastic channel dropout. More recently, RFNet \cite{ding2021rfnet} introduced region-aware fusion for handling missing modalities, and knowledge distillation approaches have been explored to transfer complete-modality knowledge to incomplete-modality students \cite{Kundu2025Distillation}.

However, these strategies have been primarily developed for CNN backbones. In Transformer-based architectures such as Swin UNETR, the interaction between modality-specific information and the self-attention mechanism has not been explicitly modeled. Our MMCF module addresses this by introducing a learnable, zero-initialized residual branch that adaptively reweights modalities before they enter the Transformer encoder, preserving the pretrained initialization while enabling progressive modality adaptation.

\subsection{Decoder optimization and deep supervision}

Deep supervision, which applies auxiliary losses at intermediate decoder stages, was introduced to improve gradient flow and preserve spatial detail \cite{Lee2015DSN}. UNet++ \cite{Zhou2018UNetpp} extended this idea through nested dense connections. In the context of brain tumor segmentation, deep supervision has been shown to improve convergence and boundary accuracy in CNN-based models.

However, in Transformer-based segmentation architectures, decoder design has received comparatively less attention. UNETR \cite{Hatamizadeh2021UNETR} and Swin UNETR \cite{Hatamizadeh2022SwinUNETR} employ powerful hierarchical encoders with relatively conventional decoder reconstruction paths. nnFormer \cite{Zhou2023nnFormer} introduced interleaved convolution and self-attention in the decoder, leaving input-conditioned decoder regularization as an open design opportunity.

Our complete DDS design extends standard deep supervision in two ways: (1) it gates decoder features using the modality-aware attention map from MMCF, creating an information pathway from input-level modality analysis to decoder-level reconstruction; and (2) it uses a residual gating mechanism initialized near identity, preserving decoder function when the attention signal is uninformative. In the configuration comparison, the Aux. supervision row reports the decoder-supervision setting without MMCF-derived gating, whereas the complete MMCF+DDS configuration evaluates the gated variant.

\subsection{Uncertainty estimation in medical image segmentation}

Reliable deployment of segmentation models requires accurate predictions accompanied by calibrated confidence estimates. Bayesian neural networks provide a principled framework for epistemic uncertainty by placing distributions over model parameters \cite{Blundell2015BayesByBackprop}. Practical approximations include Monte Carlo (MC) dropout \cite{Gal2016DropoutBayesian}, which interprets dropout at test time as approximate variational inference, and deep ensembles \cite{Lakshminarayanan2017Ensemble}, which estimate uncertainty through prediction disagreement across independently trained models.

For 3D medical image segmentation, these approaches face computational challenges. MC dropout requires multiple forward passes (typically 10--50), multiplying inference time. Deep ensembles require training multiple full models, multiplying training cost. For large Transformer architectures with 60+ million parameters, both approaches become expensive. Recent work has explored selective Bayesianization, where only critical layers are made probabilistic \cite{Alawad2024ProbBNN}, and batch ensembles \cite{Jain2023BatchEnsemble}, which share most parameters across ensemble members.

Our approach follows the selective Bayesianization philosophy: we convert only the final convolutional layer (which maps decoder features to class probabilities) to a Bayesian layer after deterministic training has converged. This design choice is motivated by the observation that the final layer is where class-boundary decisions are made, and uncertainty in these decisions is most directly relevant to segmentation reliability. By initializing the Bayesian posterior from converged deterministic weights, we achieve rapid fine-tuning convergence (30 epochs vs. 150 for the deterministic stage).

\subsection{Summary and positioning}

The above review reveals that modality-aware fusion, decoder regularization, and uncertainty estimation have been studied largely as separate research threads. MyoPS-Net \cite{qiu2023myops} demonstrated the value of flexible multi-sequence combination for cardiac pathology segmentation, and DFuse-Net \cite{zhou2024multi} showed that disentangled representation learning with uncertainty analysis improves brain tumor segmentation reliability. However, no existing work combines adaptive modality fusion, decoder-level regularization, and efficient Bayesian inference within a single Transformer-based framework for brain tumor segmentation. BMDS-Net addresses this gap through a modular, two-stage design that enables controlled configuration-level comparisons.

\section{Method}
\label{sec3}

\subsection{Problem formulation and design rationale}

\begin{figure}[!t]
    \centering
    \includegraphics[width=0.9\textwidth]{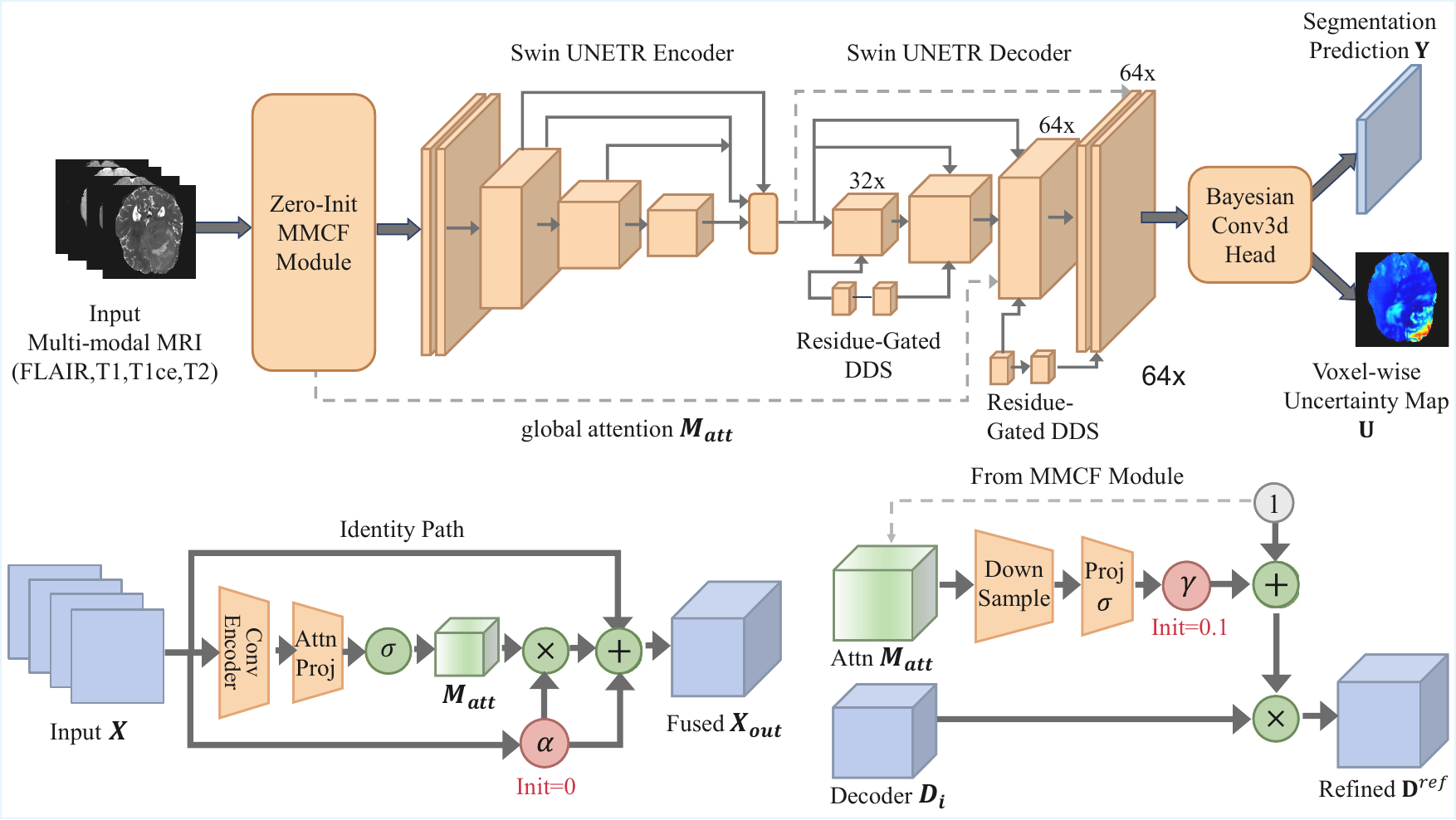}
    \Description{Architecture diagram of BMDS-Net showing MMCF before a shared Swin UNETR encoder, residual-gated decoder supervision, and Bayesian output-layer fine-tuning.}
    \caption{Architecture of BMDS-Net. \textbf{Top:} Overall pipeline showing the two-stage training strategy. The input passes through MMCF for adaptive modality fusion, then enters the Swin UNETR backbone. The attention map $\mathbf{M}_{att}$ is propagated to the decoder for deep supervision via DDS. \textbf{Bottom left:} MMCF adaptively reweights modalities through a zero-initialized residual branch, ensuring stable initialization. \textbf{Bottom right:} DDS uses $\mathbf{M}_{att}$ to gate decoder features $\mathbf{D}_i$ through a residual modulation pathway, regularizing boundary reconstruction.}
    \label{fig:framework}
\end{figure}

Let $\mathcal{D} = \{(\mathbf{X}_n, \mathbf{Y}_n)\}_{n=1}^N$ denote a multi-modal MRI dataset, where $\mathbf{X}_n \in \mathbb{R}^{C_{in} \times H \times W \times D}$ is a 3D volume with $C_{in}=4$ modalities (FLAIR, T1, T1ce, T2), and $\mathbf{Y}_n \in \{0,1\}^{C_{out} \times H \times W \times D}$ denotes voxel-wise labels for $C_{out}=3$ overlapping BraTS evaluation regions (WT, TC, and ET). The network therefore produces three independent region probabilities, and non-overlapping visual labels (NCR/NET, ED, ET) are derived from the nested WT/TC/ET predictions for display. The goal is to learn a mapping:
\begin{equation}
\Phi: \mathbb{R}^{C_{in} \times \Omega} \rightarrow [0,1]^{C_{out} \times \Omega}, \quad \Omega = H \times W \times D.
\end{equation}

From a probabilistic perspective, segmentation approximates the posterior $p(\mathbf{Y}|\mathbf{X})$. A deterministic network estimates a point estimate of parameters $\theta^*$ and produces $p(\mathbf{Y}|\mathbf{X}, \theta^*)$. A Bayesian approach instead marginalizes over parameter uncertainty: $p(\mathbf{Y}|\mathbf{X}) = \int p(\mathbf{Y}|\mathbf{X}, \theta) p(\theta|\mathcal{D}) d\theta$. Our two-stage design approximates this: Stage~1 finds a strong point estimate $\theta^*$ for the bulk of the network, and Stage~2 introduces a variational posterior over the final layer to capture residual predictive uncertainty.

The overall architecture is shown in Figure~\ref{fig:framework}. We chose Swin UNETR as the backbone because it provides hierarchical feature extraction through shifted-window self-attention, achieving strong baseline performance on BraTS while maintaining a manageable parameter count. Our augmentations (MMCF, DDS, Bayesian layer) are designed to be modular: each addresses a specific limitation without disrupting the backbone's learned representations.

\subsection{Stage 1: Deterministic learning with modality-aware fusion and decoder regularization}

\subsubsection{Zero-initialized Multimodal Contextual Fusion (MMCF)}

The design of MMCF is motivated by two observations. First, standard early concatenation treats all modalities as equally informative, which is suboptimal when modalities carry different signal-to-noise ratios or when a modality is absent. Second, naively adding a fusion module before a pretrained encoder can disrupt the encoder's learned representations if the fusion module produces large initial perturbations.

To address both issues, MMCF employs a zero-initialized residual design. Given input $\mathbf{X} \in \mathbb{R}^{C_{in} \times H \times W \times D}$, a lightweight convolutional encoder $\mathcal{F}_{enc}$ (two 3D convolutional layers with group normalization) extracts intermediate features, and an attention head $\mathcal{C}_{att}$ (1$\times$1$\times$1 convolution followed by sigmoid) produces a spatial-channel attention map:
\begin{equation}
\mathbf{F}_{feat} = \mathcal{F}_{enc}(\mathbf{X}), \quad
\mathbf{M}_{att} = \sigma(\mathcal{C}_{att}(\mathbf{F}_{feat})) \in [0,1]^{C_{in} \times H \times W \times D}
\end{equation}
where $\sigma$ denotes the sigmoid function. The fused input is computed as:
\begin{equation}
\mathbf{X}_{fused} = \mathbf{X} + \alpha \cdot (\mathbf{X} \odot \mathbf{M}_{att})
\end{equation}
where $\alpha \in \mathbb{R}$ is a learnable scalar \textbf{initialized to zero}. This initialization ensures that at the start of training, $\mathbf{X}_{fused} = \mathbf{X}$ exactly, preserving the Swin UNETR pretrained initialization. As training progresses, $\alpha$ controls the strength of a modality-aware residual contribution, allowing the network to emphasize reliable modality evidence without abruptly perturbing the input distribution. This progressive adaptation avoids the optimization instability that can arise from abrupt input perturbation, analogous to the zero-initialization strategy used in adapter modules for large language models.

\textbf{Behavior under missing modalities.} When a modality channel is zeroed out (simulating clinical absence), the corresponding entries in $\mathbf{M}_{att}$ receive no informative input. Because the attention is applied multiplicatively ($\mathbf{X} \odot \mathbf{M}_{att}$), the zeroed channel contributes nothing to the residual branch regardless of the attention value. The remaining modalities, which do carry signal, can still contribute to the residual pathway through their attention values. This provides a soft, learned form of modality-aware adaptation without requiring explicit missing-modality training.

\subsubsection{Residual-Gated Deep Decoder Supervision (DDS)}

Standard deep supervision applies auxiliary classification heads at intermediate decoder stages to improve gradient flow. However, these auxiliary heads operate independently of the input-level modality analysis. Our DDS module creates an explicit information pathway from MMCF's attention map to the decoder, encouraging decoder features to be consistent with the modality-aware context identified at the input.

Let $\mathbf{D}_i$ denote the feature map at decoder stage $i$. We resize $\mathbf{M}_{att}$ to the resolution of $\mathbf{D}_i$ via trilinear interpolation, then apply a projection layer $\mathcal{P}_{proj}$ (1$\times$1$\times$1 convolution) to produce a decoder gate:
\begin{equation}
\mathbf{G}_i = 1 + \gamma \cdot \sigma(\mathcal{P}_{proj}(\text{Interp}(\mathbf{M}_{att})))
\end{equation}
\begin{equation}
\mathbf{D}^{refined}_i = \mathbf{D}_i \odot \mathbf{G}_i
\end{equation}
where $\gamma$ is a learnable scaling factor initialized to 0.1. The ``$1 + \ldots$'' formulation ensures that the gate operates as a residual modulation: when $\gamma$ is small, $\mathbf{G}_i \approx 1$ and the decoder features pass through almost unchanged. As training progresses, the gate learns to allocate stronger residual emphasis to regions where the input attention is high.

\textbf{Input-attention-guided decoder gating.} Using the input-level attention map $\mathbf{M}_{att}$ creates a consistency constraint: the decoder is encouraged to reconstruct features that align with the modality reliability pattern identified at the input. This acts as a form of feature distillation from encoder to decoder, which we formalize through the feature consistency loss below.

\subsubsection{Training objective for Stage 1}

The deterministic stage is optimized with a composite objective combining segmentation supervision and feature consistency regularization.

\textbf{BoundaryFocalDice loss.} The principal segmentation objective is a project-level composite loss that combines Dice overlap, channel-wise binary focal loss, and a boundary-restricted binary cross-entropy term:
\begin{equation}
    \mathcal{L}_{BFD} = \alpha_d \mathcal{L}_{Dice} + \alpha_b \mathcal{L}_{Boundary} + \alpha_f \mathcal{L}_{Focal}.
\end{equation}
Unless otherwise stated, we use $\alpha_d=\alpha_b=\alpha_f=1$ and focal exponent $\eta=2$. For compact notation, let $C=C_{fg}=C_{out}=3$ denote the three overlapping BraTS foreground regions. For predicted sigmoid probabilities $\mathbf{P}$ and three-channel multi-hot region masks $\mathbf{Y}$, the Dice term is
\begin{equation}
    \mathcal{L}_{Dice}=1-\frac{1}{C}\sum_{c=1}^{C}\frac{2\sum_v P_{c}(v)Y_{c}(v)+\epsilon}{\sum_v P_{c}(v)+\sum_v Y_{c}(v)+\epsilon}.
\end{equation}
The binary focal term is computed channel-wise from binary cross-entropy:
\begin{equation}
    \ell_{BCE}(v,c)=-Y_c(v)\log P_c(v)-[1-Y_c(v)]\log[1-P_c(v)],
\end{equation}
\begin{equation}
    \mathcal{L}_{Focal}=\frac{1}{C|\Omega|}\sum_{c=1}^{C}\sum_{v\in\Omega}(1-p_{t,c}(v))^{\eta}\ell_{BCE}(v,c),
\end{equation}
where $p_{t,c}(v)=P_c(v)$ when $Y_c(v)=1$ and $p_{t,c}(v)=1-P_c(v)$ otherwise. For each tumor region, a binary boundary mask $\mathcal{B}_c$ is extracted from the corresponding ground-truth region mask using a 3D Sobel gradient operator. The gradient magnitude is thresholded at 0.5 to form a one-voxel boundary support without additional dilation. If a region has no boundary voxels in a crop, $\epsilon$ keeps the denominator nonzero and the numerator contributes zero. The boundary term averages binary cross-entropy only over these boundary voxels:
\begin{equation}
    \mathcal{L}_{Boundary}=\sum_{c=1}^{C_{fg}}\frac{\sum_{v\in\Omega}\mathcal{B}_c(v)\,\ell_{BCE}(v,c)}{\sum_{v\in\Omega}\mathcal{B}_c(v)+\epsilon}.
\end{equation}
Thus, the implemented boundary term sums the class-wise boundary BCE averages. In code, the same term is evaluated with the numerically stable logit form $\ell_{BCEWithLogits}(Z_c,Y_c)=\operatorname{softplus}(Z_c)-Y_cZ_c$, which is equivalent to applying BCE to $P_c=\sigma(Z_c)$. This loss is used to emphasize difficult voxels and tumor interfaces without changing the inference graph.

\textbf{Segmentation loss.} Auxiliary supervision is applied at two intermediate decoder outputs with spatial sizes $32^3$ and $64^3$ (for $128^3$ input crops) using the same BoundaryFocalDice objective and decreasing weights:
\begin{equation}
    \mathcal{L}_{seg} = \mathcal{L}_{BFD}(\mathbf{Y}_{final}, \mathbf{Y}_{gt}) + 0.4 \cdot \mathcal{L}_{BFD}(\mathbf{Y}_{aux}^{(1)}, \mathbf{Y}_{gt}) + 0.2 \cdot \mathcal{L}_{BFD}(\mathbf{Y}_{aux}^{(2)}, \mathbf{Y}_{gt})
\end{equation}
The weight schedule ($\lambda_1=0.4$, $\lambda_2=0.2$) follows the common practice of assigning smaller weights to shallower auxiliary outputs \cite{Zhou2018UNetpp}, reflecting the intuition that coarser predictions should contribute less to the total gradient.

\textbf{Feature consistency loss.} To align encoder-derived attention with decoder reconstruction, we first average the modality-channel attention map into a spatial saliency map, $\bar{\mathbf{M}}_{att}$. We then minimize the discrepancy between the spatial activation pattern of refined decoder features and the interpolated saliency map:
\begin{equation}
    \mathcal{L}_{distill} = \sum_{i} \left\| \mathcal{N}(\|\mathbf{D}^{refined}_i\|_2) - \mathcal{N}(\text{Interp}(\bar{\mathbf{M}}_{att})) \right\|^2_2
\end{equation}
where $\bar{\mathbf{M}}_{att}=\frac{1}{C_{in}}\sum_{c=1}^{C_{in}}\mathbf{M}_{att}^{(c)}$, $\|\cdot\|_2$ denotes channel-wise L2 norm, and $\mathcal{N}(\cdot)$ denotes min-max spatial normalization. This loss encourages the decoder to allocate reconstruction effort proportionally to the modality-aware saliency identified at the input.

The total Stage~1 objective is:
\begin{equation}
\mathcal{L}_{total} = \mathcal{L}_{seg} + 0.05 \cdot \mathcal{L}_{distill}
\end{equation}

\subsection{Stage 2: Efficient Bayesian fine-tuning for uncertainty estimation}

\subsubsection{Motivation and design choice}

The deterministic model from Stage~1 produces accurate segmentations but provides no measure of predictive confidence. Full Bayesian treatment of the entire network (62M+ parameters) is computationally prohibitive. We therefore adopt a \emph{last-layer Bayesian} strategy, motivated by two considerations:

\begin{enumerate}
    \item The final convolutional layer is where the network makes its class-boundary decisions. Uncertainty in this layer directly translates to uncertainty in the segmentation output.
    \item By initializing the Bayesian posterior from converged deterministic weights, we leverage the strong feature representations learned in Stage~1 and only need to learn the \emph{residual uncertainty} around these features.
\end{enumerate}

\subsubsection{Variationally regularized fine-tuning}

The final 1$\times$1$\times$1 convolutional layer is replaced by a \textbf{BayesianConv3d} layer whose weights $\mathcal{W}$ follow a variational posterior $q_\theta(\mathcal{W}) = \mathcal{N}(\mu, \sigma^2)$. The posterior mean $\mu$ is initialized from the converged deterministic weights, providing a warm start. The variance is parameterized via the softplus function: $\sigma = \log(1 + \exp(\rho))$, with $\rho$ initialized to produce small initial variance.

Weights are sampled using the reparameterization trick:
\begin{equation}
    \mathcal{W} = \mu + \log(1 + \exp(\rho)) \odot \epsilon, \quad \epsilon \sim \mathcal{N}(0, \mathbf{I})
\end{equation}

Bayesian fine-tuning uses a KL-regularized variational composite segmentation objective:
\begin{equation}
    \mathcal{L}_{var} = \mathcal{L}_{BFD}(\Phi(\mathbf{X}; \mathcal{W}), \mathbf{Y}) + \beta_{KL} \cdot D_{KL}(q_\theta(\mathcal{W}) \| p(\mathcal{W}))
\end{equation}
where $p(\mathcal{W}) = \mathcal{N}(0, \mathbf{I})$ is a standard Gaussian prior and $\beta_{KL}$ controls the regularization strength. Because the data term is the composite BoundaryFocalDice objective rather than a pure negative log-likelihood, we interpret this objective as variational regularization of the final layer. We use a single Monte Carlo sample ($T=1$) during training for computational efficiency.

\subsubsection{Uncertainty quantification at inference}

During inference, we perform $T=20$ stochastic forward passes with different weight samples. Predictive uncertainty is quantified voxel-wise as the class-averaged variance of sigmoid region probabilities:
\begin{equation}
    \sigma^2_{pred}(v) = \frac{1}{C_{out}}\sum_{c=1}^{C_{out}}\frac{1}{T} \sum_{t=1}^{T} (\hat{p}_{t,c}(v) - \bar{p}_{c}(v))^2
\end{equation}
where $\hat{p}_{t,c}(v)$ is the predicted probability for region channel $c$ at voxel $v$ from the $t$-th stochastic pass and $\bar{p}_{c}(v) = \frac{1}{T}\sum_t \hat{p}_{t,c}(v)$ is the mean prediction. High variance indicates regions where the final Bayesian layer is uncertain given the fixed deterministic feature representation and may be elevated near tumor boundaries or heterogeneous tissue interfaces.

The complete two-stage training procedure is summarized in Algorithm~\ref{alg1}.

\begin{algorithm}[!t]
\caption{Two-stage training procedure for BMDS-Net}
\label{alg1}
\begin{algorithmic}[1]
\REQUIRE Multi-modal dataset $\mathcal{D} = \{(x_i, y_i)\}$, Stage~1 epochs $E_1$, Stage~2 epochs $E_2$
\ENSURE Trained Bayesian model $\mathcal{M}_{Bayes}$
\STATE \textbf{Stage 1: Deterministic optimization}
\STATE Initialize Swin UNETR with pretrained weights $\theta_{det}$
\STATE Initialize MMCF scalar $\alpha \leftarrow 0$, DDS scalar $\gamma \leftarrow 0.1$
\FOR{epoch $e = 1$ \TO $E_1$}
    \FOR{each batch $(x, y) \in \mathcal{D}$}
        \STATE $x_{fused}, \mathbf{M}_{att} \leftarrow \text{MMCF}(x)$
        \STATE $z \leftarrow \text{Encoder}(x_{fused})$
        \STATE $D_{refined} \leftarrow \text{DDS}(\text{Decoder}(z), \mathbf{M}_{att})$
        \STATE Compute $\mathcal{L}_{total} = \mathcal{L}_{seg} + 0.05 \cdot \mathcal{L}_{distill}$
        \STATE Update $\theta_{det}, \alpha, \gamma$ via AdamW
    \ENDFOR
\ENDFOR
\STATE \textbf{Stage 2: Bayesian fine-tuning}
\STATE Replace final Conv3d with BayesianConv3d
\STATE Initialize posterior mean $\mu \leftarrow \theta_{det}^{\text{final}}$, variance $\rho \leftarrow$ small
\STATE Freeze all parameters except Bayesian layer
\FOR{epoch $e = 1$ \TO $E_2$}
    \FOR{each batch $(x, y) \in \mathcal{D}$}
        \STATE Sample $\mathcal{W} \sim q(\mathcal{W}|\mu, \rho)$ via reparameterization
        \STATE Compute $\mathcal{L}_{var} = \mathcal{L}_{BFD} + \beta_{KL} \cdot D_{KL}(q \| p)$
        \STATE Update $\mu, \rho$ via AdamW
    \ENDFOR
\ENDFOR
\RETURN $\mathcal{M}_{Bayes}$
\end{algorithmic}
\end{algorithm}

\section{Experiments and Results}
\label{sec4}

\subsection{Dataset and experimental protocol}

\textbf{Dataset.} The primary experiments used the BraTS 2021 dataset \cite{Baid2021BraTSBenchmark}, which contains 1,251 multi-institutional, multi-modal MRI cases with expert-annotated voxel-wise labels for three hierarchical tumor sub-regions: whole tumor (WT = edema + necrosis + enhancing), tumor core (TC = necrosis + enhancing), and enhancing tumor (ET). Each case includes four co-registered MRI sequences (FLAIR, T1, T1ce, T2) at 1~mm isotropic resolution. The multi-institutional nature of BraTS introduces natural variability in acquisition protocols, scanner manufacturers, and image quality, providing a realistic evaluation setting. To evaluate whether the internal configuration ranking persisted across a separate public BraTS release, we additionally used a BraTS 2020 supervised cohort \cite{Menze2015BRATS,Bakas2017SciData} with the same four-sequence input format and tumor sub-region definitions.

\textbf{Data split and preprocessing.} BraTS 2021 data were split into training (80\%, 1,000 cases), validation (15\%, 188 cases), and held-out test (5\%, 63 cases) sets with stratification by tumor volume. Unless otherwise stated, reported comparisons, configuration analyses, robustness tests, and calibration analyses refer to the BraTS 2021 validation cohort. The BraTS 2020 same-training experiment used 294 training cases and 74 validation cases, with the five internal configurations trained and evaluated under a shared protocol. Each modality was independently z-score normalized. Training used random crops of $128 \times 128 \times 128$ voxels with online augmentation including random rotation ($\pm$15\textdegree), elastic deformation, and intensity scaling.

\textbf{Implementation details.} BMDS-Net was implemented in PyTorch 1.12 with MONAI 1.0. Stage~1 was trained for 150 epochs using AdamW (weight decay $10^{-4}$) with cosine annealing warm restarts (initial learning rate $10^{-4}$, minimum $10^{-6}$). Stage~2 fine-tuned the Bayesian layer for 30 epochs with a reduced learning rate ($10^{-5}$). The KL weight $\beta_{KL}$ was set to $10^{-6}$ following a warm-up schedule over the first 5 epochs. All experiments were conducted on NVIDIA RTX 3090 GPUs (24~GB).

\textbf{Evaluation metrics.} Segmentation quality was assessed using Dice Similarity Coefficient (DSC) and 95th-percentile Hausdorff Distance (HD95). Main segmentation tables report whole-volume validation metrics after reconstructing each case from sliding-window inference. Calibration quality was assessed using Expected Calibration Error (ECE) and Negative Log-Likelihood (NLL) over the three WT/TC/ET sigmoid output channels, excluding the background label. ECE was computed with 15 equal-width confidence bins after pooling voxel-channel predictions across validation cases; NLL was computed as the mean three-channel binary cross-entropy. Robustness was assessed by evaluating performance under systematic single-modality removal and additive Gaussian noise.

\textbf{Controlled internal configurations.} To keep the configuration naming consistent across tables and figures, we use the following definitions. Baseline denotes the original Swin UNETR training configuration. Loss only keeps the Swin UNETR inference graph unchanged and replaces the baseline objective with BoundaryFocalDice, without auxiliary heads, MMCF, decoder gating, or distillation. Aux. supervision builds on the BoundaryFocalDice objective and adds auxiliary decoder heads at the two intermediate resolutions, but removes MMCF and MMCF-derived decoder gating at inference. MMCF only uses BoundaryFocalDice with the zero-initialized MMCF input-fusion branch, without auxiliary decoder heads or gated DDS. BMDS-Net combines BoundaryFocalDice, MMCF residual fusion, auxiliary decoder supervision, gated DDS, and the MMCF--DDS feature consistency loss.

This protocol creates a staged evidence ladder. BraTS 2021 establishes the main comparison, configuration, robustness, and calibration results; BraTS 2020 then tests whether the internal ranking of the five Swin UNETR-controlled configurations remains stable under a separate public BraTS cohort trained with the same recipe.

\subsection{Comparison with representative methods}

We compared BMDS-Net with six representative methods spanning CNN-based (SegResNet, MedNeXt, nnU-Net), hybrid (TransBTS), and Transformer-based (nnFormer, Swin UNETR) architectures. For methods marked with *, results were obtained from recent benchmarking studies \cite{Wu2025FCFDiff, Hatamizadeh2022UNetFormer} using the same dataset. All other results were obtained by training with identical data splits and preprocessing.

\textbf{Analysis of results.} As shown in Table~\ref{tab:sota}, BMDS-Net achieved the highest WT Dice (0.9293) and the lowest HD95 for both WT (2.27~mm) and TC (2.22~mm). The HD95 improvements over Swin UNETR are most visible for TC and ET: TC improved from 2.39 to 2.22~mm and ET from 3.84 to 3.27~mm, while WT changed from 2.30 to 2.27~mm. Because HD95 measures a high-percentile boundary distance, these results indicate improved near-worst-case boundary localization, with the strongest practical signal in TC and ET.

For Dice scores, the improvements over Swin UNETR are modest (WT: +0.14\%, ET: +0.46\%), and nnU-Net achieved higher TC Dice (0.9193 vs. 0.9098). This pattern is expected: at Dice levels above 0.90, further improvements become increasingly difficult due to annotation noise and inherent ambiguity in tumor boundaries. The more informative comparison is in HD95, where BMDS-Net shows consistent improvements across all three regions, suggesting that the DDS module successfully regularizes boundary reconstruction.

BMDS-Net achieves these improvements with only 2\% additional parameters (63.27M vs. 61.99M) over Swin UNETR, indicating that the added fusion and decoder pathways provide boundary benefits with limited model growth.

\begin{table}[H]
  \caption{Comparison with representative methods on the BraTS 2021 validation set. WT, whole tumor; TC, tumor core; ET, enhancing tumor. Best values are in bold. The gray row marks the strongest direct Transformer baseline, and the blue row marks BMDS-Net. The final row reports the change relative to Swin UNETR; green indicates improvement and red indicates additional cost. * indicates results from benchmarking studies.}
  \label{tab:sota}
  \centering
  \small
  \setlength{\tabcolsep}{3.4pt}
  \renewcommand{\arraystretch}{1.08}
  \begin{tabularx}{\textwidth}{@{}lYYYYYYYY@{}}
    \toprule
    \textbf{Method} & \multicolumn{3}{c}{\textbf{Dice Score $\uparrow$}} & \multicolumn{3}{c}{\textbf{HD95 (mm) $\downarrow$}} & \textbf{Params} & \textbf{FLOPs} \\
    \cmidrule(lr){2-4} \cmidrule(lr){5-7}
     & \textbf{WT} & \textbf{TC} & \textbf{ET} & \textbf{WT} & \textbf{TC} & \textbf{ET} & \textbf{(M)} & \textbf{(G)} \\
    \midrule
    SegResNet \cite{Myronenko2018AutoencoderReg} & 0.9053 & 0.8814 & 0.8578 & 11.97 & 9.73 & 3.77 & \textbf{4.70} & \textbf{148.04} \\
    MedNeXt \cite{Roy2023MedNeXt} & 0.9187 & 0.8882 & 0.8426 & 6.18 & 5.36 & 4.41 & 7.91 & 218.22 \\
    TransBTS* \cite{Wang2021TransBTS} & 0.9230 & 0.8810 & 0.8480 & 4.23 & 4.39 & \textbf{3.16} & 32.99 & 333.00 \\
    nnFormer* \cite{Zhou2023nnFormer} & 0.9268 & 0.9015 & \textbf{0.8687} & 4.43 & 8.04 & 3.27 & 149.0 & 106.5 \\
    nnU-Net \cite{Isensee2021nnUNet} & 0.9292 & \textbf{0.9193} & 0.8672 & 4.87 & 3.92 & 3.51 & 30.75 & 412.50 \\
    \rowcolor{tabgray} Swin UNETR \cite{Hatamizadeh2022SwinUNETR} & 0.9279 & 0.9111 & 0.8629 & 2.30 & 2.39 & 3.84 & 61.99 & 794.02 \\
    \midrule
    \rowcolor{tabblue} \oursmethod{BMDS-Net (Ours)} & \best{0.9293} & 0.9098 & 0.8675 & \best{2.27} & \best{2.22} & 3.27 & 63.27 & 872.26 \\
    \multicolumn{1}{l}{\neutral{$\Delta$ vs. Swin UNETR}} & \gain{+0.0014} & \drop{-0.0013} & \gain{+0.0046} & \gain{-0.03} & \gain{-0.17} & \gain{-0.57} & \drop{+1.28} & \drop{+78.24} \\
    \bottomrule
  \end{tabularx}
\end{table}

\FloatBarrier

\begin{figure}[H]
  \centering
  \includegraphics[width=0.86\textwidth]{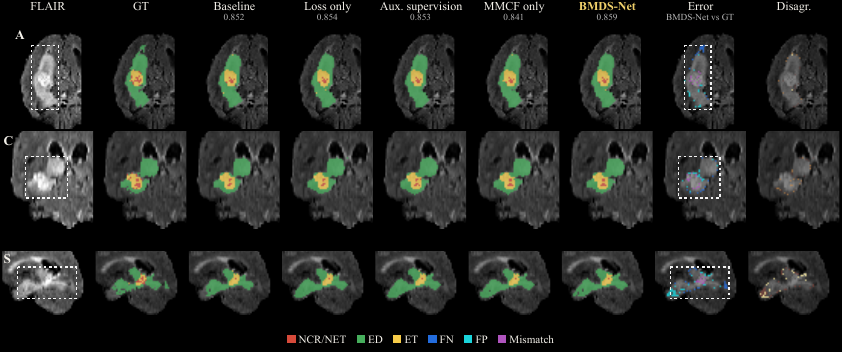}
  \Description{Qualitative BraTS 2021 comparison showing FLAIR input, ground truth, internal configuration predictions, BMDS-Net error map, and method-disagreement proxy.}
  \caption{Qualitative comparison on validation case BraTS2021\_00320. The plate shows the input FLAIR image, ground truth, internal variants, BMDS-Net, the error map, and a method-disagreement proxy. Error and disagreement panels summarize localized false-positive, false-negative, and inter-configuration boundary variation; values above prediction panels denote case-level mean Dice.}
  \label{fig:qualitative}
\end{figure}

\textbf{Qualitative analysis.} Figure~\ref{fig:qualitative} complements the aggregate metrics by showing how each internal configuration changes the same tumor region in a validation case. The baseline captures the dominant enhancing tumor but leaves scattered boundary errors. The loss-only variant keeps the Swin UNETR inference graph and replaces the training objective with the BoundaryFocalDice loss; the auxiliary-supervision variant further adds decoder supervision without MMCF-derived gating. Across the displayed variants, local discrepancies change most visibly around tumor interfaces, motivating the joint use of overlap metrics, error maps, and uncertainty summaries.

\FloatBarrier

\subsection{Computational efficiency}

Table~\ref{tab:efficiency} reports inference latency while separating external architecture baselines from Swin UNETR-controlled configurations. Measurements use a $128 \times 128 \times 128$ input crop. Auxiliary heads are removed at test time; the 0.93~ms difference between Swin UNETR and the Aux. supervision row is within timing variation for the unchanged inference graph. The gated DDS pathway in the complete BMDS-Net remains active and adds 2.07~ms and 12.26~G FLOPs relative to MMCF-only inference. The Bayesian mode with $T=20$ passes requires $\sim$4.1~seconds per $128^3$ crop, positioning uncertainty estimation as an offline review mode, while deterministic BMDS-Net remains the segmentation mode for routine mask generation.

\begin{table}[H]
  \caption{Inference efficiency on NVIDIA RTX 3090 with input size $128 \times 128 \times 128$. Results separate architecture-level baselines, backbone-controlled configurations, and uncertainty-enabled inference. Auxiliary heads are removed at test time; gated DDS remains active in deterministic BMDS-Net, and the Bayesian stage uses $T=20$ stochastic forward passes.}
  \label{tab:efficiency}
  \centering
  \scriptsize
  \setlength{\tabcolsep}{3.0pt}
  \renewcommand{\arraystretch}{0.96}
  \begin{tabularx}{\textwidth}{@{}lYYY@{}}
    \toprule
    \textbf{Configuration} & \textbf{Type} & \textbf{Latency (ms) $\downarrow$} & \textbf{FPS $\uparrow$} \\
    \midrule
    \multicolumn{4}{@{}l}{\tabnote{\textit{Architecture-level baselines}}} \\
    \addlinespace[0.1em]
    SegResNet & CNN & 58.09 & 17.21 \\
    UNETR & Transformer & 68.79 & 14.54 \\
    MedNeXt-B & CNN & 85.90 & 11.64 \\
    \midrule
    \multicolumn{4}{@{}l}{\tabnote{\textit{Backbone-controlled variants}}} \\
    \addlinespace[0.1em]
    \rowcolor{tabgray} Swin UNETR (Baseline) & Transformer & 187.27 & 5.34 \\
    MMCF only & Transformer & 202.55 & 4.94 \\
    Aux. supervision & Transformer & 188.20 & 5.31 \\
    \rowcolor{tabblue} \oursmethod{BMDS-Net (deterministic)} & Transformer & 204.62 & 4.89 \\
    \midrule
    \multicolumn{4}{@{}l}{\tabnote{\textit{Uncertainty-enabled inference}}} \\
    \addlinespace[0.1em]
    \rowcolor{tabblue} \oursmethod{BMDS-Net (Bayesian, $T$=20)} & Transformer & $\sim$4092 & 0.24 \\
    \midrule
    \multicolumn{2}{@{}l}{\neutral{$\Delta$ deterministic vs. Swin UNETR}} & \drop{+17.35} & \drop{-0.45} \\
    \bottomrule
  \end{tabularx}
\end{table}

\FloatBarrier

\subsection{Configuration comparison: understanding component behaviour}

\begin{table}[H]
  \caption{Configuration comparison on the BraTS 2021 validation set. Rows are backbone-controlled configurations evaluated on the full validation cohort. The gray row marks the baseline and the blue row marks the complete BMDS-Net. The Aux. supervision row evaluates the BoundaryFocalDice-trained decoder-supervision configuration without the MMCF input-fusion residual or MMCF-derived decoder gating; these training-only auxiliary heads are disabled during inference, so the reported parameter and FLOP counts match the inference graph. The complete MMCF+DDS row evaluates the gated DDS variant, whose projection and residual-gating pathway remain active at inference.}
  \label{tab:ablation}
  \centering
  \small
  \setlength{\tabcolsep}{3.5pt}
  \renewcommand{\arraystretch}{1.08}
  \begin{tabularx}{\textwidth}{@{}lYYYYYYYY@{}}
    \toprule
    \textbf{Configuration} & \multicolumn{3}{c}{\textbf{Dice Score $\uparrow$}} & \multicolumn{3}{c}{\textbf{HD95 (mm) $\downarrow$}} & \textbf{Params} & \textbf{FLOPs} \\
    \cmidrule(lr){2-4} \cmidrule(lr){5-7}
     & \textbf{WT} & \textbf{TC} & \textbf{ET} & \textbf{WT} & \textbf{TC} & \textbf{ET} & \textbf{(M)} & \textbf{(G)} \\
    \midrule
    \rowcolor{tabgray} Baseline (Swin UNETR) & 0.9279 & 0.9111 & 0.8629 & 2.30 & 2.39 & 3.84 & \textbf{61.99} & \textbf{794.02} \\
    MMCF only & 0.9288 & 0.9095 & 0.8654 & 2.21 & 2.44 & 3.32 & 62.03 & 860.00 \\
    Aux. supervision & \textbf{0.9312} & \textbf{0.9144} & \textbf{0.8718} & \textbf{2.10} & \textbf{1.93} & \textbf{2.83} & 61.99 & 794.02 \\
    \rowcolor{tabblue} \oursmethod{MMCF + DDS (BMDS-Net)} & 0.9293 & 0.9098 & 0.8675 & 2.27 & 2.22 & 3.27 & 63.27 & 872.26 \\
    \midrule
    \multicolumn{1}{l}{\neutral{$\Delta$ BMDS-Net vs. baseline}} & \gain{+0.0014} & \drop{-0.0013} & \gain{+0.0046} & \gain{-0.03} & \gain{-0.17} & \gain{-0.57} & \drop{+1.28} & \drop{+78.24} \\
    \bottomrule
  \end{tabularx}
\end{table}

The configuration results in Table~\ref{tab:ablation} separate complete-input optimization from deployment-oriented robustness. On BraTS 2021, the Aux. supervision configuration provides the strongest complete-input metrics without changing the inference graph, while the full BMDS-Net combines MMCF residual fusion with gated DDS and is evaluated for its robustness profile in Table~\ref{tab:robustness}. The BraTS 2020 experiment in Table~\ref{tab:brats2020_ablation} provides the fuller five-configuration ladder, including the Loss only setting.

\textbf{Interpretation.} The complete-input profile of the Aux. supervision configuration is consistent with the role of additional decoder losses: they provide extra gradient pathways and regularize decoder reconstruction. Under complete inputs, all modalities carry useful information, and the MMCF attention map is relatively uniform. In this regime, MMCF adds parameters and computation while offering limited additional signal for complete-input overlap. Under missing-modality evaluation, MMCF-only improves some conditions but degrades others; the robust pattern appears when MMCF residual fusion and gated DDS are combined, where BMDS-Net achieves the best mean Dice under every single-modality removal setting.

This condition-dependent behavior ties configuration choice to the expected operating regime. Where complete modalities are consistently available, the auxiliary-supervision setting provides a compact complete-input configuration. Where modality absence is plausible, the full BMDS-Net provides stronger worst-case behaviour.

\subsection{Second-dataset validation on BraTS 2020}

\begin{figure}[H]
  \centering
  \includegraphics[width=\textwidth]{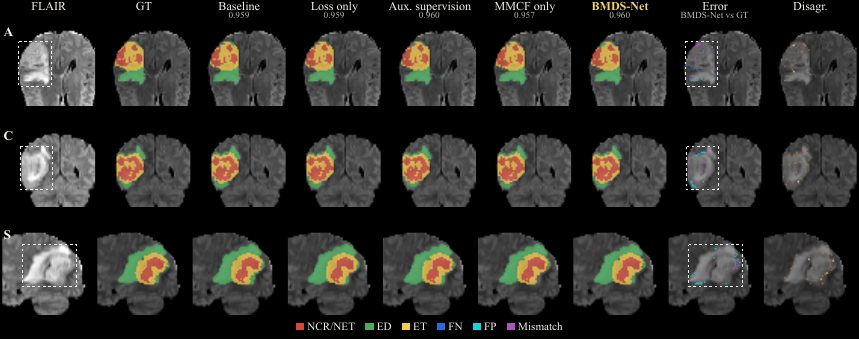}
  \Description{BraTS 2020 qualitative comparison showing FLAIR input, ground truth, five internal configuration predictions, BMDS-Net error map, and method disagreement.}
  \caption{BraTS 2020 qualitative internal configuration comparison on validation case BraTS20\_Training\_359. The plate follows the same visual protocol as Figure~\ref{fig:qualitative}: FLAIR image, ground truth, five internal configurations, BMDS-Net error map, and method disagreement. Residual disagreement remains concentrated near small tumor-interface structures, while the displayed case illustrates the TC/ET boundary behaviour observed in the cohort-level comparison.}
  \label{fig:brats2020_qualitative}
\end{figure}

Table~\ref{tab:brats2020_ablation} extends the internal configuration comparison to BraTS 2020. The complete BMDS-Net achieved the highest mean Dice (0.8229) and led the tumor-core and enhancing-tumor Dice scores among the five controlled configurations. The metric profile was region-specific: WT Dice remained effectively tied with the baseline, while the full model improved the average score, TC/ET overlap, and WT/TC boundary distance. Figure~\ref{fig:brats2020_qualitative} visualizes the same five-configuration protocol on a BraTS 2020 validation case. Together, the quantitative and qualitative BraTS 2020 results reinforce the mechanism-level reading from BraTS 2021, where the complete configuration contributes most consistently when accuracy is evaluated together with boundary behaviour and deployment-relevant robustness.

\begin{table}[H]
  \caption{Internal configuration comparison on the BraTS 2020 validation cohort (74 cases) using the same training protocol. WT, whole tumor; TC, tumor core; ET, enhancing tumor. Best values are in bold. The gray row marks the Swin UNETR baseline and the blue row marks the complete BMDS-Net.}
  \label{tab:brats2020_ablation}
  \centering
  \small
  \setlength{\tabcolsep}{3.3pt}
  \renewcommand{\arraystretch}{1.08}
  \begin{tabularx}{\textwidth}{@{}lYYYYYYY@{}}
    \toprule
    \textbf{Configuration} & \multicolumn{4}{c}{\textbf{Dice Score $\uparrow$}} & \multicolumn{3}{c}{\textbf{HD95 (mm) $\downarrow$}} \\
    \cmidrule(lr){2-5} \cmidrule(lr){6-8}
     & \textbf{WT} & \textbf{TC} & \textbf{ET} & \textbf{Mean} & \textbf{WT} & \textbf{TC} & \textbf{ET} \\
    \midrule
    \rowcolor{tabgray} Baseline (Swin UNETR) & \best{0.8976} & 0.8435 & 0.7158 & 0.8189 & 8.04 & 10.48 & \best{8.63} \\
    Loss only & 0.8967 & 0.8329 & 0.7193 & 0.8163 & 7.37 & 10.71 & 10.42 \\
    Aux. supervision & 0.8971 & 0.8352 & 0.7171 & 0.8165 & 8.73 & 11.10 & 9.47 \\
    MMCF only & 0.8601 & 0.7526 & 0.7147 & 0.7758 & 45.52 & 55.71 & 8.69 \\
    \rowcolor{tabblue} \oursmethod{MMCF + DDS (BMDS-Net)} & 0.8969 & \best{0.8443} & \best{0.7273} & \best{0.8229} & \best{7.15} & \best{9.21} & 8.83 \\
    \bottomrule
  \end{tabularx}
\end{table}

\subsection{Hyperparameter sensitivity and regional sensitivity analysis}

\begin{table}[H]
  \caption{Sensitivity of segmentation performance to the initialization value of the MMCF fusion parameter $\alpha$. The blue row marks the proposed zero initialization. Values report final WT Dice from the initialization-sensitivity evaluation. The zero-initialized setting matches the best non-zero initialization ($\alpha=1.0$) at the reported precision.}
  \label{tab:sensitivity}
  \centering
  \small
  \setlength{\tabcolsep}{3.4pt}
  \renewcommand{\arraystretch}{1.08}
  \begin{tabularx}{0.62\textwidth}{@{}YY@{}}
    \toprule
    \textbf{$\alpha$ Initialization} & \textbf{WT Dice $\uparrow$} \\
    \midrule
    \rowcolor{tabblue} \oursmethod{0 (proposed)} & \best{0.9293} \\
    0.5 & 0.9255 \\
    \rowcolor{tabgray} 1.0 & \best{0.9293} \\
    1.5 & 0.9281 \\
    2.0 & 0.9262 \\
    \bottomrule
  \end{tabularx}
\end{table}

Table~\ref{tab:sensitivity} shows the sensitivity of WT Dice to the initialization of $\alpha$. The zero-initialization strategy matches the best non-zero initialization ($\alpha=1.0$), while non-optimal initializations ($\alpha=0.5, 2.0$) lead to slightly lower WT Dice. These results indicate that zero initialization did not reduce final WT Dice at the reported precision and preserves the pretrained backbone's learned representations at the start of training.

\begin{figure}[H]
  \centering
  \includegraphics[width=0.82\linewidth]{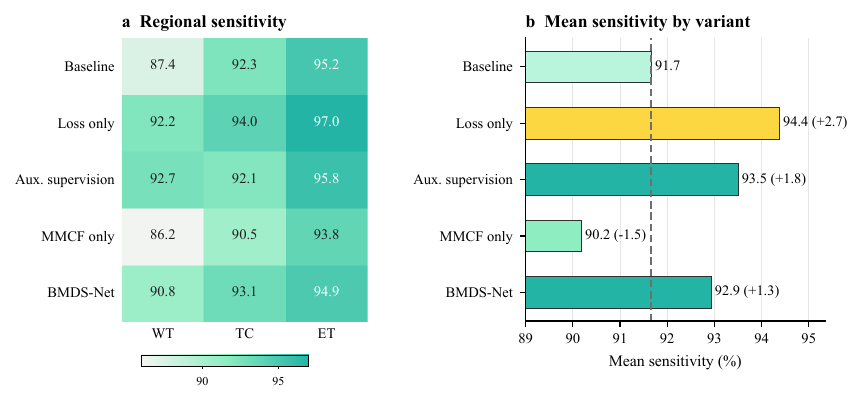}
  \Description{Regional sensitivity figure with a heatmap for WT, TC, and ET sensitivity and a zoomed mean-sensitivity bar plot across internal variants.}
  \caption{Regional sensitivity analysis for internal variants on BraTS2021\_00320. \textbf{a,} Case-level sensitivity for WT, TC, and ET. \textbf{b,} Mean sensitivity across variants on a zoomed bar axis, reported together with the change relative to the baseline. Higher sensitivity reflects fewer false negatives and is interpreted alongside Dice because aggressive over-segmentation can increase recall while reducing precision.}
  \label{fig:regional_sensitivity}
\end{figure}

Figure~\ref{fig:regional_sensitivity} separates sensitivity from Dice to distinguish recall from overall segmentation quality. In this case, Loss only and Aux. supervision achieved the highest mean sensitivity (94.4\% and 93.5\%, respectively), whereas BMDS-Net achieved 92.9\%. This pattern indicates that the case-level Dice profile was not driven solely by higher recall, but by a trade-off between missed tumor, false positives, and boundary localization.

\FloatBarrier

\subsection{Robustness to missing modalities}

\begin{table}[H]
  \caption{Robustness evaluation under missing-modality and noise perturbations. Values are mean Dice (standard deviation) averaged across WT, TC, and ET. The gray row marks the baseline and the blue row marks BMDS-Net. The final row reports the change relative to the baseline. Models were trained with complete modalities and evaluated with systematic single-modality removal (zero-filling) at test time.}
  \label{tab:robustness}
  \centering
  \small
  \setlength{\tabcolsep}{3.6pt}
  \renewcommand{\arraystretch}{1.08}
  \begin{tabularx}{\textwidth}{@{}lYYYYYY@{}}
    \toprule
    \textbf{Configuration} & \textbf{Full} & \textbf{Missing T1} & \textbf{Missing T1ce} & \textbf{Missing T2} & \textbf{Missing FLAIR} & \textbf{Noise (0.1)} \\
    \midrule
    \rowcolor{tabgray} Baseline & 0.901 (0.110) & 0.768 (0.218) & 0.848 (0.152) & 0.364 (0.100) & 0.877 (0.120) & 0.901 (0.110) \\
    MMCF only & 0.901 (0.110) & 0.752 (0.218) & 0.855 (0.145) & 0.353 (0.109) & 0.879 (0.122) & 0.901 (0.110) \\
    Aux. supervision & \textbf{0.906} (0.103) & 0.768 (0.207) & 0.865 (0.139) & 0.369 (0.116) & 0.880 (0.114) & \textbf{0.906} (0.102) \\
    \rowcolor{tabblue} \oursmethod{BMDS-Net} & 0.902 (0.106) & \textbf{0.783} (0.209) & \textbf{0.868} (0.137) & \textbf{0.388} (0.115) & \textbf{0.881} (0.123) & 0.901 (0.111) \\
    \midrule
    \multicolumn{1}{l}{\neutral{$\Delta$ BMDS-Net vs. baseline}} & \gain{+0.001} & \gain{+0.015} & \gain{+0.020} & \gain{+0.024} & \gain{+0.004} & \neutral{+0.000} \\
    \bottomrule
  \end{tabularx}
\end{table}

Table~\ref{tab:robustness} presents the central robustness analysis. Several observations emerge:

\textbf{Missing T2 causes the largest degradation across all models.} Performance drops from $\sim$0.90 to $\sim$0.36--0.39. In this model and evaluation protocol, T2 carried the most indispensable edema-related information, as indicated by the largest performance degradation after T2 removal.

\textbf{BMDS-Net provides the best robustness under all missing-modality conditions.} Compared to the baseline, BMDS-Net improves mean Dice by +1.5\% (missing T1), +2.0\% (missing T1ce), +2.4\% (missing T2), and +0.4\% (missing FLAIR). The largest improvements occur for missing T1ce and T2 under this removal protocol, indicating that the model benefits most from adaptive fusion when the omitted sequence carries distinctive information for the evaluated regions.

\textbf{MMCF and DDS contribute complementarily in the full model.} Auxiliary supervision improves full-data performance and is also stronger than MMCF-only in several missing-modality settings. MMCF-only improves missing T1ce and missing FLAIR but degrades missing T1 and missing T2. The complete BMDS-Net combines residual fusion with gated DDS and achieves the best robustness under all four single-modality removal settings while maintaining near-optimal full-data performance.

\textbf{Limited degradation under tested noise.} Gaussian noise with std = 0.1 produced no measurable degradation at the reported precision, suggesting that this perturbation level did not materially affect the evaluated Swin UNETR-based models.

\FloatBarrier

\subsection{Uncertainty estimation and calibration}

\begin{table}[H]
  \caption{Comparison of uncertainty estimation approaches. Training cost is relative to one deterministic model. The gray row marks the deterministic Swin UNETR reference and the blue row marks Bayesian BMDS-Net. ECE, expected calibration error; NLL, negative log-likelihood. Lower ECE and NLL indicate better calibration. Because both the deterministic architecture and the output-layer treatment differ between the first and third rows, the comparison is interpreted at the system level.}
  \label{tab:uncertainty_comparison}
  \centering
  \small
  \setlength{\tabcolsep}{3.4pt}
  \renewcommand{\arraystretch}{1.08}
  \begin{tabularx}{\textwidth}{@{}lYYY@{}}
    \toprule
    \textbf{Method} & \textbf{Training Cost} & \textbf{ECE $\downarrow$} & \textbf{NLL $\downarrow$} \\
    \midrule
    \rowcolor{tabgray} Deterministic (Swin UNETR) & $1\times$ & 0.0152 & 0.0421 \\
    Deep Ensemble ($N=3$) & $3\times$ & \textbf{0.0035} & \textbf{0.0210} \\
    \rowcolor{tabblue} \oursmethod{BMDS-Net (Bayesian)} & $\approx 1.2\times$ & 0.0037 & 0.0218 \\
    \midrule
    \multicolumn{1}{l}{\neutral{$\Delta$ BMDS-Net vs. deterministic}} & \drop{+0.2$\times$} & \gain{-0.0115} & \gain{-0.0203} \\
    \bottomrule
  \end{tabularx}
\end{table}

Table~\ref{tab:uncertainty_comparison} compares calibration performance against two references. The deterministic Swin UNETR baseline exhibits substantial overconfidence (ECE = 0.0152), a well-known property of deep networks trained with cross-entropy loss \cite{Guo2017Calibration}. The three-model deep ensemble reduces ECE to 0.0035 but requires $3\times$ training cost ($\sim$450 GPU-hours for Swin UNETR). Bayesian BMDS-Net achieves similar calibration (ECE = 0.0037) at $\sim$1.2$\times$ training cost, corresponding to approximately 2.5$\times$ lower cost than the ensemble. This system-level comparison reflects the combined effect of the deterministic BMDS architecture and the Bayesian output-layer treatment.

\begin{figure}[H]
  \centering
  \includegraphics[width=\linewidth]{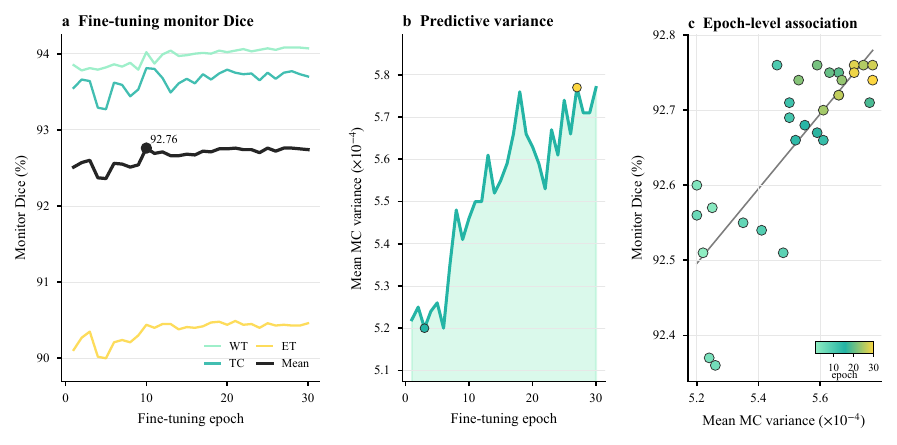}
  \Description{Uncertainty analysis plots showing Bayesian fine-tuning Dice, Monte Carlo predictive variance, and their epoch-level association.}
  \caption{Uncertainty analysis during Bayesian fine-tuning. \textbf{a,} Region-wise and mean Dice across 30 Bayesian fine-tuning epochs under the internal monitoring metric used for optimization. \textbf{b,} Mean Monte Carlo predictive variance, used here as an epistemic uncertainty summary. \textbf{c,} Descriptive epoch-level association between mean predictive variance and mean Dice during optimization.}
  \label{fig:uncertainty}
\end{figure}

Figure~\ref{fig:uncertainty} reports the internal monitoring trajectory used during Bayesian fine-tuning. This monitor is the epoch-level validation Dice logged by the fine-tuning loop on $128^3$ validation samples with batch size 1, averaged over the WT, TC, and ET channels after thresholded sigmoid predictions; the main segmentation table uses whole-volume validation Dice. Under this optimization monitor, mean Dice increases from 92.51\% to a stable plateau around 92.7\%, while the mean Monte Carlo variance remains in a narrow range ($5.2$--$5.8 \times 10^{-4}$). The trajectory characterizes fine-tuning stability, and the calibration table provides the system-level confidence summary. Together, these results support stable last-layer Bayesian adaptation, with voxel-level clinical review depending on spatial uncertainty--error analysis.

\textbf{Summary of experimental evidence.} Across the validation experiments, the evidence supports a deployment-specific claim: the auxiliary-supervision configuration provides the strongest complete-input segmentation and boundary regularization, the complete MMCF+gated-DDS configuration provides the most consistent missing-modality robustness, and Bayesian BMDS-Net achieves system-level calibration close to an ensemble with limited additional training cost. The BraTS 2020 same-training experiment preserves the leading mean-Dice ranking of the complete BMDS-Net among five internal configurations, while also showing that the gain is concentrated in TC/ET overlap and WT/TC boundary distance. This pattern positions the full model as a deployment-aware configuration whose value emerges from the coordinated behaviour of segmentation accuracy, robustness, and calibration.

\section{Discussion}
\label{sec5}

\subsection{Condition-dependent configuration behaviour}

A central finding of this work is that configuration behaviour is condition-dependent. Under complete-modality conditions, the auxiliary-supervision configuration shows the clearest complete-input advantage, consistent with direct regularization of the decoder reconstruction pathway and improved boundary metrics (HD95). MMCF, in contrast, provides a relatively uniform attention map when all modalities carry useful information, offering limited additional signal.

Under missing-modality conditions, the main bottleneck shifts toward degraded input evidence. Auxiliary decoder supervision remains useful, and MMCF alone shows mixed behaviour across different missing sequences; the complete MMCF+gated-DDS configuration provides the best worst-case performance across all tested missing-modality conditions.

This finding has practical implications for deployment. In clinical settings where complete modalities are reliably available (e.g., dedicated neuro-oncology centers with standardized protocols), the auxiliary-supervision variant may be preferred for its slightly better full-data performance and lower computational cost. In settings where modality absence is common (e.g., retrospective studies, emergency imaging, resource-limited facilities), the full BMDS-Net showed better empirical robustness under the evaluated removal settings.

\subsection{Why last-layer Bayesian inference is effective}

Our results show that the BMDS-Net system with a last-layer Bayesian head achieves calibration (ECE = 0.0037) close to a three-model ensemble (ECE = 0.0035). This is notable given that only the final prediction layer is made probabilistic. We interpret this result through two considerations:

First, the final layer is where class-boundary decisions are made. With the encoder and decoder fixed during Bayesian fine-tuning, the uncertainty estimate reflects residual decision uncertainty conditioned on the deterministic feature representation.

Second, the deterministic Stage~1 training already produces well-separated feature representations, as reflected by high Dice scores and low HD95 values. After this deterministic convergence, a useful fraction of the residual uncertainty is likely concentrated near class boundaries and ambiguous tissue interfaces. A single Bayesian decision layer can therefore provide a compact uncertainty summary without requiring variational inference over the entire 3D Transformer.

This interpretation is consistent with the fine-tuning dynamics in Figure~\ref{fig:uncertainty}: the Bayesian fine-tuning metric remains stable while the system-level calibration table improves relative to the deterministic Swin UNETR reference. A natural next step is to quantify the spatial overlap between uncertainty, boundary regions, and segmentation errors, thereby connecting calibration summaries to voxel-level review.

\subsection{The role of zero initialization}

The zero-initialization strategy for MMCF's $\alpha$ parameter serves a dual purpose. Technically, it ensures that the pretrained Swin UNETR backbone receives unmodified inputs at the start of training, preserving its learned representations. Conceptually, it implements a form of ``progressive complexity'': the model first learns to segment with standard concatenation (leveraging pretrained weights), then gradually discovers beneficial modality reweighting as $\alpha$ grows from zero.

The sensitivity analysis (Table~\ref{tab:sensitivity}) indicates that this strategy achieves the same final WT Dice as the optimal non-zero initialization ($\alpha=1.0$), while avoiding the need to search for the optimal initial value. This is analogous to the zero-initialization strategy used in LoRA adapters for large language models, where new parameters are initialized to produce zero output, ensuring that the adapted model starts from the pretrained solution.

\subsection{Failure modes and limitations}

Despite the improvements demonstrated, several failure modes persist:

\textbf{Post-operative cavities vs. necrotic core.} In patients with prior surgical resection, the post-operative cavity can appear similar to necrotic tumor core on T1ce, leading to false-positive TC predictions. This ambiguity sits at the image-appearance level, so temporal information (pre- vs. post-operative scans) or explicit cavity modeling would be needed to separate surgical change from viable tumor.

\textbf{Irrecoverable information loss under missing T2.} All models show severe degradation when T2 is absent (Dice $\sim$0.36--0.39), indicating that the evaluated models relied strongly on T2 for edema-related segmentation evidence under this protocol. This failure mode points toward modality synthesis or imputation as a complementary direction when edema-sensitive contrast is unavailable.

\textbf{Cohort scope.} BraTS 2021 provides the primary development and validation setting, and BraTS 2020 provides a second public BraTS cohort for same-protocol internal-configuration consistency. Both cohorts share harmonized BraTS annotation conventions; institutional cohorts with different scanner protocols, treatment histories, and clinical workflows would further define the transportability of the framework.

\textbf{Bayesian inference latency.} The $T=20$ stochastic passes required for uncertainty estimation increase inference time by $20\times$ compared to deterministic mode. For time-critical applications, this could be mitigated by reducing $T$ (with some calibration degradation) or by distilling the uncertainty estimates into a deterministic auxiliary head.

\textbf{Missing-modality training strategy.} Our robustness evaluation tests models trained with complete modalities on incomplete inputs at test time. Explicit modality dropout augmentation could further improve robustness and would provide a complementary axis for studying how architectural adaptation and training-time modality perturbation interact.

\subsection{Clinical implications}

The practical value of BMDS-Net is concentrated in three deployment-relevant properties:

\begin{enumerate}
    \item \textbf{Graceful degradation.} When a modality is unavailable, BMDS-Net degrades more gracefully than the baseline, retaining higher segmentation performance under incomplete imaging.
    \item \textbf{Boundary reliability.} TC and ET HD95 improvements suggest more reliable high-percentile boundary localization for contour review.
    \item \textbf{Calibration-aware review.} The Bayesian BMDS-Net system achieves substantially lower ECE than the deterministic Swin UNETR reference, supporting future human-in-the-loop workflows in which uncertainty maps could prioritize regions for review after spatial validation.
\end{enumerate}

These properties collectively support a deployment model where BMDS-Net provides initial segmentations together with calibrated confidence summaries. The next translational step is evaluation on institutional cohorts and prospective reader studies that measure how calibrated uncertainty changes review time, boundary correction, and downstream planning decisions.

\section{Conclusion}
\label{sec6}

We presented BMDS-Net, a two-stage framework that turns Swin UNETR into a deployment-aware segmentation system through adaptive fusion, boundary-aware decoder regularization, and efficient Bayesian calibration. The evidence supports three main conclusions:

\begin{itemize}
    \item \textbf{Configuration behaviour is condition-dependent.} Auxiliary-supervision configurations favour complete-input boundary metrics, while missing-modality robustness is strongest when MMCF residual fusion and gated DDS are combined.
    \item \textbf{The full configuration is deployment-oriented.} BMDS-Net provides the best worst-case robustness on BraTS 2021 and preserves the leading mean-Dice ranking among five internal variants on BraTS 2020, with the strongest TC and ET Dice.
    \item \textbf{Calibration can be added efficiently.} Bayesian BMDS-Net approaches three-model ensemble calibration at approximately 40\% of the ensemble training budget.
\end{itemize}

These results argue for matching architectural design to the expected operating conditions: complete-modality segmentation, incomplete-modality robustness, and calibrated review.

\textbf{Future directions.} Priority areas for future work include: (1) validation on institutional multi-center cohorts outside the public BraTS ecosystem; (2) integration of explicit modality dropout during training to further improve missing-modality robustness; (3) development of faster uncertainty inference through deterministic uncertainty distillation; and (4) prospective clinical evaluation of the uncertainty-guided review workflow.

\section*{Data availability}
The BraTS 2021 and BraTS 2020 datasets are publicly available through the BraTS/Synapse challenge portals (\url{https://www.synapse.org/}). Source code and trained-model release information are available at \url{https://github.com/RyanZhou168/BMDS-Net}.

\section*{Declaration of competing interest}
The authors declare that they have no known competing financial interests or personal relationships that could have appeared to influence the work reported in this paper.

\bibliographystyle{ACM-Reference-Format}
\bibliography{paper}

\end{document}